\documentclass[letterpaper, 10 pt, conference]{ieeeconf}  

\IEEEoverridecommandlockouts                              

\usepackage{url}
\usepackage{cite}
\usepackage{times}
\usepackage{graphicx}
\usepackage{amsmath}
\usepackage{amssymb}
\usepackage{enumerate}
\usepackage{mathtools}
\usepackage{amsmath}
\usepackage{algorithm}  
\usepackage{algorithmic}

\usepackage{subfigure}
\usepackage{float}

\title{\LARGE \bf
Real-time Scalable Dense Surfel Mapping
}

\author{Kaixuan Wang, Fei Gao and Shaojie Shen
\thanks{All authors are with the Department of Electronic and Computer Engineering, The Hong Kong University of Science and Technology, Hong Kong, SAR China. {\tt\small \{kwangap, fgaoaa, eeshaojie\}@ust.hk}}
}

\begin{document}

\maketitle
\thispagestyle{empty} 
\pagestyle{empty}



\begin{abstract}

In this paper, we propose a novel dense surfel mapping system that scales well in different environments with only CPU computation. Using a sparse SLAM system to estimate camera poses, the proposed mapping system can fuse intensity images and depth images into a globally consistent model. The system is carefully designed so that it can build from room-scale environments to urban-scale environments using depth images from RGB-D cameras, stereo cameras or even a monocular camera. First, superpixels extracted from both intensity and depth images are used to model surfels in the system. superpixel-based surfels make our method both run-time efficient and memory efficient. Second, surfels are further organized according to the pose graph of the SLAM system to achieve $O(1)$ fusion time regardless of the scale of reconstructed models. Third, a fast map deformation using the optimized pose graph enables the map to achieve global consistency in real-time. The proposed surfel mapping system is compared with other state-of-the-art methods on synthetic datasets. The performances of urban-scale and room-scale reconstruction are demonstrated using the KITTI dataset~\cite{kitti} and autonomous aggressive flights, respectively. The code is available for the benefit of the community\footnote{https://github.com/HKUST-Aerial-Robotics/DenseSurfelMapping}.


\end{abstract}

\section{INTRODUCTION}

Estimating the surrounding 3D environment is one of the fundamental abilities for robots to navigate safely or operate high-level tasks. To be usable in mobile robot applications, the mapping system needs to fulfill the following four requirements. First, the 3D reconstruction has to densely cover the environment in order to provide sufficient information for navigation. Second, the mapping system should have good scalability and efficiency so that it can be deployed in different environments using limited onboard computation resources. From room-scale (several meters) to urban-scale (several kilometers) environments, the mapping system should maintain both run-time efficiency and memory efficiency. Third, global consistency is required in the mapping systems. If loops are detected, the system should be able to deform the map in real-time to maintain consistency between different visits. Fourth, to be usable in different robot applications, the system should be able to fuse depth maps of different qualities from RGB-D cameras, stereo cameras or even monocular cameras.

\begin{figure}[h]
\begin{center}
\includegraphics[width=0.95\linewidth]{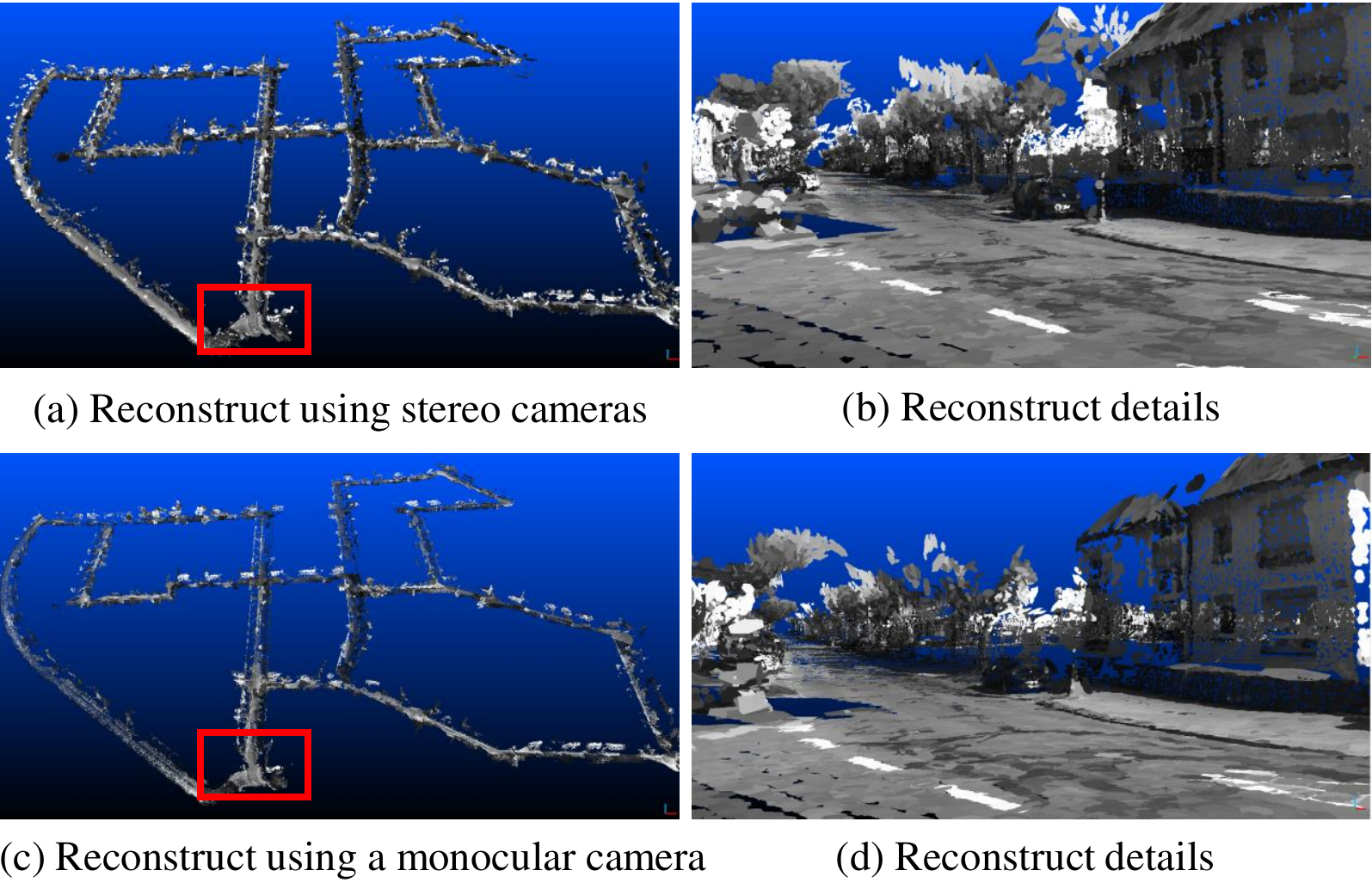}
\end{center}
\vspace*{-0.4cm}
\caption{Our proposed dense mapping method can fuse low-quality depth maps to reconstruct large-scale globally-consistent environments in real-time without GPU acceleration. The top row shows the reconstruction of \textit{KITTI odometry 00} using stereo cameras and the detail of a looped corner. The bottom row shows the reconstruction using only one monocular camera with depth prediction~\cite{Godard_2017_CVPR}.}
\vspace*{-0.9cm}
\label{fig:kitti_reconstruction}
\end{figure}

In recent years, many methods have been proposed to reconstruct the environment using RGB-D cameras focusing on several requirements mentioned above. KinectFusion~\cite{izadi2011kinectfusion} is a pioneering work that uses the truncated signed distance field (TSDF)~\cite{tsdf} to represent 3D environments. Many following works improve the scalability (e.g. Kintinuous~\cite{kintinuous}), the efficiency (e.g. CHISEL~\cite{openchisel}), and the global consistency (e.g. BundleFusion~\cite{bundlefusion}) of TSDF-based methods. Surfel-based methods model the environment as a collection of surfels. For example, ElasticFusion~\cite{elastic_fusion} uses surfels to reconstruct the scene and achieves global consistency. Although all these methods achieve impressive results using RGB-D cameras, extending them to fulfill all four requirements and to be usable in different robot applications is non-trivial.

In this paper, we propose a mapping method that fulfills all four requirements and can be applied to a range of mobile robotic systems. Our system uses state-of-the-art sparse visual SLAM systems to track camera poses and fuses intensity images and depth images into a globally consistent model. Unlike ElasticFusion~\cite{elastic_fusion} that treats each pixel as a surfel, we use superpixels to represent surfels. Pixels are clustered into superpixels if they share similar intensity, depth, and spatial locations. Modeling superpixels as surfels greatly reduces the memory requirement of our system and enables the system to fuse noisy depth maps from stereo cameras or a monocular camera. Surfels are organized according to the keyframes they are last observed in. Using the pose graph of the SLAM systems, we further find locally consistent keyframes and surfels that the relative drift between each other is negligible. Only locally consistent surfels are fused with input images, achieving $O(1)$ fusion time and local accuracy. Global consistency is achieved by deforming surfels according to the optimized pose graph. Thanks to the careful design, our system can be used to reconstruct globally consistent urban-scale environments in real-time without GPU acceleration.

In summary, the main contributions of our mapping method are the following.
\begin{itemize}
\item We use superpixels extracted from both intensity and depth images to model surfels in the system. Superpixels enable our method to fuse low-quality depth maps. Run-time efficiency and memory efficiency are also gained by using superpixel-based surfels.
\item We further organize surfels accordingly to the pose graph of the sparse SLAM systems. Using this organization, locally consistent maps are extracted for fusion, and the fusion time maintains $O(1)$ regardless of the reconstruction scale. Fast map deformation is also proposed based on the optimized pose graph so that the system can achieve global consistency in real-time.
\item We implement the proposed dense mapping system using only CPU computation. We evaluate the method using public datasets and demonstrate its usability using autonomous aggressive flights. To the best of our knowledge, the proposed method is the first online depth fusion approach that achieves global consistency in urban-scale using only CPU computation.


\end{itemize}

\section{RELATED WORK}


Most online dense reconstruction methods take depth maps from RGB-D cameras as input. In this section, we introduce different methods to extend the scalability, global consistency and run-time efficiency of these mapping systems.

Kintinuous~\cite{kintinuous} extends the scalability of mapping systems by using a cyclical buffer. The TSDF volume is virtually transformed according to the movement of the camera. Voxel hashing proposed by Nie{\ss}ner et al.~\cite{niessner2013real} is another solution to improve the scalability. Due to the sparsity of the surfaces in the space, only valid voxels are stored using hashing functions. DynSLAM~\cite{tsdf_kitti} reconstructs urban-scale models using hashed voxels and a high-end GPU to accelerate. Surfel-based methods are relatively scalable compared with voxel-based methods because only surfaces are stored in the system. Without the explicit data optimization, ElasticFusion~\cite{elastic_fusion} can build room-scalable environments in detail. Fu et al.~\cite{large_scale_surfel} further increase the scalability of surfel-based methods by maintaining a local surfel set.

To remove the drift from camera tracking and maintain global consistency, mapping systems should be able to fast deform the model when loops are detected. Whelan et al.~\cite{deformation_tsdf} improved Kintinuous~\cite{kintinuous} with point clouds deformation. A deformation graph is constructed incrementally as the camera moves. When loops are detected, the deformation graph is optimized and applied to the point clouds. Surfel-based methods usually deform the map using similar methods. BundleFusion~\cite{bundlefusion} introduces another solution to achieve global consistency using de-integration and reintegration of RGB-D frames. When the camera poses are updated due to the pose graph optimization, RGB-D frames are firstly de-integrated from the TSDF volume and reintegrated using the updated camera poses. Submaps are used by many TSDF-based methods, such as InfiniTAM~\cite{infiniTAM}, to generate globally consistent results. These methods divide the space into multiple low-drift submaps and merge them into a global model using updated poses.

Different methods have been proposed to accelerate the fusion process. Steinbr\"ucker et al.~\cite{volumetric_cpu} use an octree as the data structure to represent the environment. Voxblox~\cite{oleynikova2017voxblox} is designed for planning that both TSDF and the Euclidean signed distance fields are calculated. Voxblox~\cite{oleynikova2017voxblox} proposes a grouped raycasting to speed up the integration, and a novel weighting strategy to deal with the distortion caused by large voxel sizes. FlashFusion~\cite{flashfusion} uses valid chunk selection to speed up the fusion step and achieves global consistency based on the reintegration method. Most of the surfel-based mapping systems require GPUs to render index maps for data association. MREMap~\cite{cpu_surfel} defines octree-organized voxels as surfels so that it does not need GPUs. However, the reconstructed model of MREMap is voxels instead of meshes. 

\section{SYSTEM OVERVIEW}

\begin{figure}[h]
\begin{center}
\vspace{-0.4cm}
\includegraphics[width=1.0\linewidth]{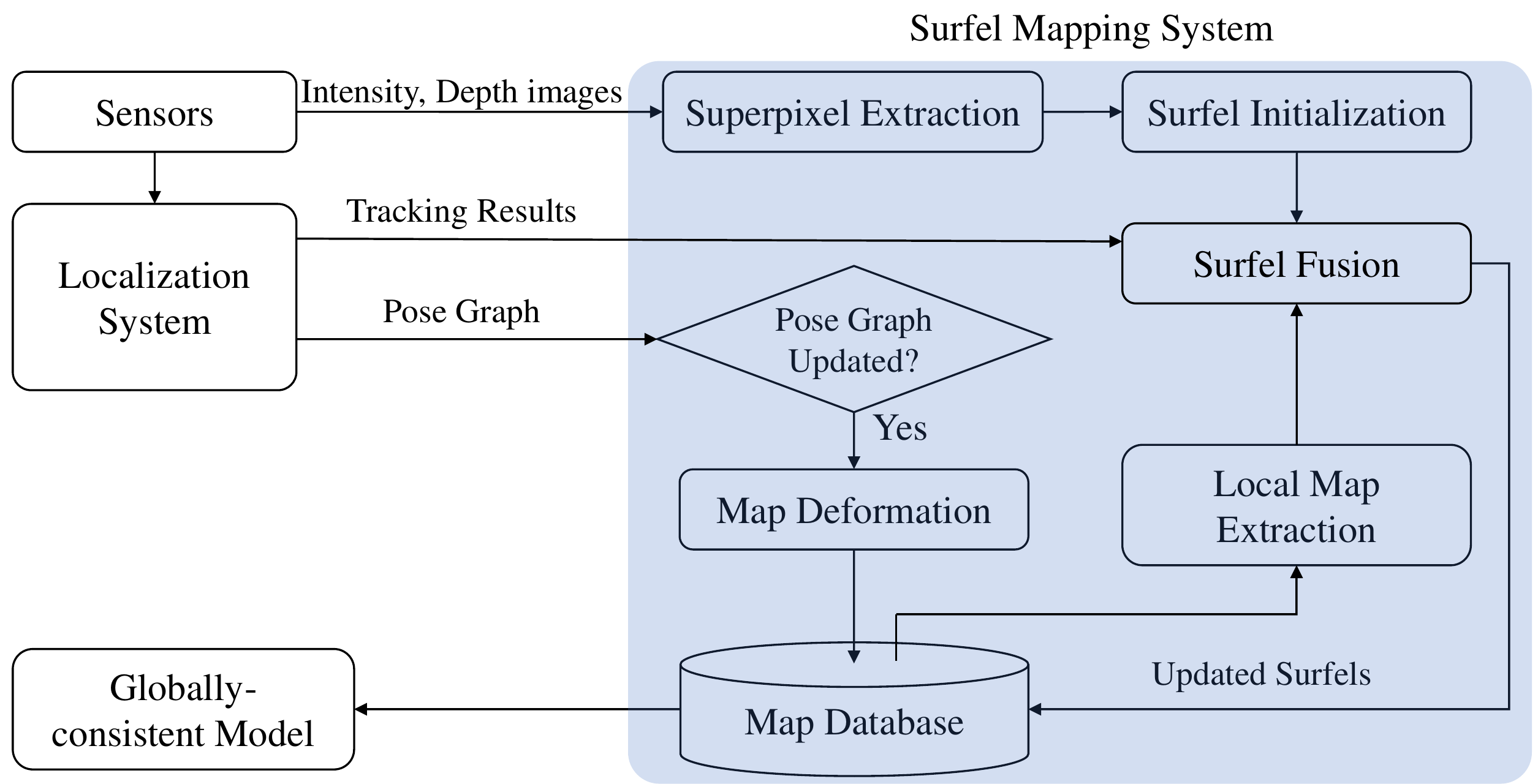}
\end{center}
\vspace{-0.4cm}
\caption{The system architecture of the proposed method. Sensors are determined by the robot application. Our system can fuse depth maps from active RGB-D cameras, stereo cameras, or monocular cameras. The localization system is used to track the camera pose, detect loops, and provide optimized pose graph of keyframes.
Each surfel in the system is attached to one keyframe, and all the surfels are stored in the map database.}
\label{fig:systemoverview}
\vspace{-0.3cm}
\end{figure}

\begin{figure*}
\begin{center}
\includegraphics[width=1.0\linewidth]{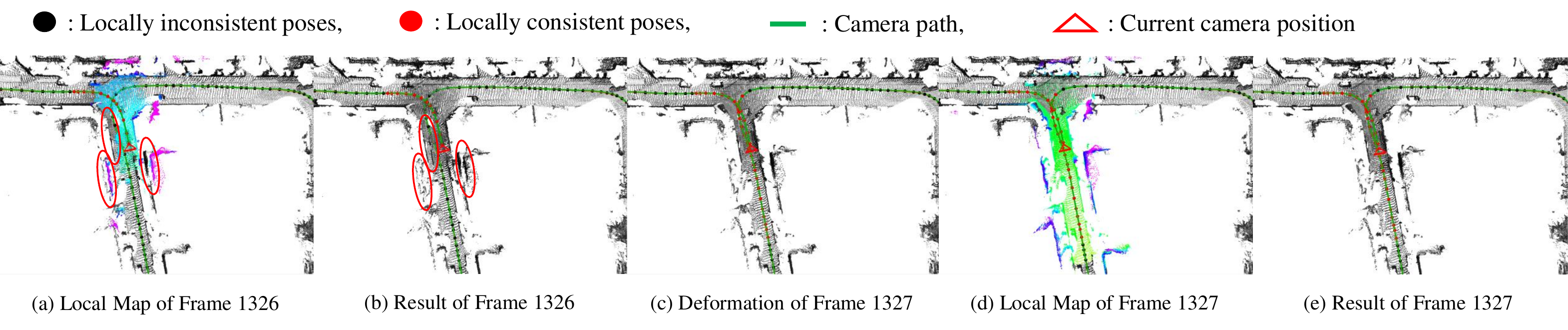}
\end{center}
\vspace{-0.5cm}
\caption{Use KITTI odometry \textit{05} as an example to show map deformation and the reuse of previous maps when the camera revisits a street. Surfels are visualized as point clouds for simplicity, and local maps are highlighted in color. During the revisit, a loop is detected on \textit{Frame 1327}. (a): local map extracted to fuse \textit{Frame 1326}. (b): the result of \textit{Frame 1326}. As highlighted in red circles in (a) and (b), the map is misaligned due to the drift before the loop closure. (c): A loop is detected by \textit{Frame 1327} and the map is deformed accordingly to remove the drift. Due to the updated pose graph, more locally consistent poses and surfels can be found. (d): Local map extracted to fuse \textit{Frame 1327}. As shown, previous maps are reused to fuse the current frame. (e) is the result after the fusion of \textit{Frame 1327}.}
\label{fig:kitti_frame}
\vspace{-0.6cm}
\end{figure*}

The system architecture is shown in Fig.~\ref{fig:systemoverview}. Our system fuses intensity and depth image pairs into a globally consistent model. We use a state-of-the-art sparse visual SLAM system (e.g. ORB-SLAM2~\cite{murORB2} or VINS-MONO~\cite{qin2017vins}) as the localization system to track the motion of the camera, detect loop closures, and optimize the pose graph. The keys to our mapping system are (1) superpixel-based surfels, (2) pose graph-based surfel fusion, and (3) fast map deformation. For each intensity and depth image input, the localization system generates camera tracking results and provides an updated pose graph. If the pose graph is optimized, our system first deforms all the surfels in the map database to ensure global consistency. After the deformation, the mapping system initializes surfels based on the extracted superpixels from the intensity and depth images. Then, local surfels are extracted from the map database according to the pose graph and fused with the initialized surfels. Finally, both the fused surfels and newly observed surfels are added back into the map database. Fig.~\ref{fig:kitti_frame} illustrates the pipeline of the system to process two frames when loops are detected.

\section{Surfel Mapping System}
\subsection{Notation}

Surfels are used to represent the environment. Each surfel $S=[S_\mathbf{p}, S_\mathbf{n}, S_c, S_w, S_r, S_t, S_i]^T$ has the following attributes: position $S_\mathbf{p} \in \mathbb{R}^3$, normal $S_\mathbf{n} \in \mathbb{R}^3$, intensity $S_c \in \mathbb{R}$, weight $S_w \in \mathbb{R}^{+}$, radius $S_r \in \mathbb{R}^{+}$, update times $S_t \in \mathbb{N}$, and the index of attached keyframe $S_i \in \mathbb{N}$. Update times $S_t$ is used to detect temporarily outliers or dynamic objects, and $S_i$ indicates the last keyframe the surfel is observed in.

Inputs of our system are intensity images, depth images, the ego-motion of the camera, and the pose graph from the SLAM system. The $i$-th intensity image is $I_i: \Omega \subset \mathbb{R}^{2} \mapsto \mathbb{R}$ and the $i$-th depth image is $D_i: \Omega \subset \mathbb{R}^{2} \mapsto \mathbb{R}$. A 3D point $\mathbf{p}=[x,y,z]^{\text{T}}$ in the camera frame can be projected into the image as a pixel $\mathbf{u} := [u,v]^{\text{T}} \in \Omega$ using the camera projection function: $\mathbf{u}=\mathbf{\pi}(\mathbf{p})$. A pixel can be back-projected into the camera frame as a point: $\mathbf{p} = \mathbf{\pi}^{-1}(\mathbf{u},d)$ where $d$ is the depth of the pixel.

\subsection{Localization System and Pose Graph}\label{localization_system}

We use a sparse visual SLAM method as the localization system to track the camera motion and optimize the pose graph when there are loop closures. For each frame, the localization system estimates the camera pose $\mathbf{T}_{w,i} \in \mathbb{SE}(3)$ and gives out the reference keyframe $F_{ref}$ that shares most features with $I_i$. $\mathbf{T}_{w,i}$ includes a rotation matrix $\mathbf{R}_{w,i} \in \mathbb{SO}(3)$ and a translation vector $\mathbf{t}_{w,i} \in \mathbb{R}^3$. Using $\mathbf{T}_{w,i}$, a point $\mathbf{p}_c$ in the camera frame of $I_i$ can be transformed into the global frame $\mathbf{p}_w = \mathbf{R}_{w,i} \mathbf{p}_c + \mathbf{t}_{w,i}$. A vector (such as the surfel normal) $\mathbf{n}_c$ in the camera frame can be transformed into the global frame $\mathbf{n}_w = \mathbf{R}_{w,i} \mathbf{n}_c$. Similiarly, $\mathbf{p}_w$ and $\mathbf{n}_w$ can be transformed back into the camera frame of $I_i$ using $\mathbf{T}_{i,w} = \mathbf{T}_{w,i}^{-1}$.

The pose graph used in our system is an undirected graph similar to the \textit{covisibility graph} in ORB-SLAM2. Vertices in the graph are the keyframes maintained in the SLAM system, and edges indicate keyframes share common features. Since the relative poses of frames are constrained by common features in the sparse SLAM systems by bundle adjustments, we assume keyframes are locally consistent if the minimum number of edges between each other is less than $G_\delta$.

\subsection{Fast Map Deformation}

If the pose graph of the localization system is updated, our method deforms all the surfels to keep the global consistency before the surfel initialization and fusion. Unlike previous methods that use a deformation graph embedded in the global map, we deform the surfels so that the relative pose between each surfel and its attached keyframe remains unchanged. Although surfels that are attached to the same keyframe are deformed rigidly, the overall deformation of the map is non-rigid.

For a surfel $S$ that is attached to keyframe $F$, the position and normal of the surfel are transformed using $\mathbf{T}_{w,\hat{F}}\mathbf{T}_{w,F}^{-1}$, where $\mathbf{T}_{w,F}$ and $\mathbf{T}_{w,\hat{F}}$ are the poses of keyframe $F$ before and after the optimization, respectively. After the deformation, the transformation ${T}_{w,F}$ is replaced by the optimized pose for the next deformation.

\subsection{Superpixel Extraction}

Unlike other surfel-based methods that model per-pixel surfels, we extract surfels based on extracted superpixels from intensity and depth images. Using superpixels greatly reduces the memory burden of our system when applied to large-scale missions. More importantly, outliers and noises from low-quality depth maps can be reduced based on extracted superpixels. This novel representation enables us to reconstruct the environment using stereo-cameras, or even monocular cameras.

Superpixels are extracted by a $k$-means approach adapted from SLIC~\cite{slic}. The original SLIC operates on RGB images and we extend it to segment both intensity and depth images. Pixels are clustered according to their intensity, depth and spatial location by firstly initializing the cluster centers and then alternating between the assignment step and the update step. A major improvement compared with SLIC is that our superpixel segmentation operates on images where not all pixels have valid depth measurements.

The cluster center $C_i = [x_i, y_i, d_i, c_i, r_i]^T$ is initialized on a regular grid on the image. $[x_i, y_i]^T$ is the average location of clustered pixels, $d_i$ is the average depth, $c_i$ is the average intensity value, and $r_i$ is the radius of the superpixel defined as the largest distance between the assigned pixels to $[x_i, y_i]^T$. $[x_i, y_i]^T$ is initialized as the location of the center. $d_i$ and $c_i$ are initialized as the depth and intensity value of pixel $[x_i, y_i]^T$. For cluster centers that are initialized on pixels with no valid depth estimations, the depth $d_i$ is initialized as NaN.

In the assignment step, the per-cluster scan from SLIC is replaced by the per-pixel update so that invalid depth can be handled while the complexity remains unchanged. We defined two distances between one pixel $\mathbf{u}$ and one candidate cluster center $C_i$ as
\begin{equation}\label{eq:distence_no_depth}
    D = \dfrac{(x_i - \mathbf{u}_x)^2+(y_i - \mathbf{u}_y)^2}{N_s^2} +  \dfrac{(c_i - \mathbf{u}_i)^2}{N_c^2},
\end{equation}
\begin{equation}\label{eq:distence_depth}
    D_{d} = D + \dfrac{(1/d_i - 1/\mathbf{u}_d)^2}{N_d^2},
\end{equation}
where $D_{d}$ and $D$ are the distances with and without depth information, respectively. [$\mathbf{u}_x$, $\mathbf{u}_y$], $\mathbf{u}_d$ and $\mathbf{u}_i$ are the location, depth and intensity of pixel $\mathbf{u}$, respectively. $N_s^2$, $N_c^2$ and $N_d^2$ are used to normalize the distance, color and depth proximity, respectively, before the summation. Each pixel scans the four neighbor candidate cluster centers. If pixel $\mathbf{u}$ and all the centers have valid depth values, then the assignment is done by comparing $D_{d}$. Otherwise, $D$ is used for the assignment.

Once all pixels have been assigned, the cluster centers are updated. $x_i$, $y_i$, and $c_i$ are updated by the average of all the assigned pixels. The mean depth $d_i$, on the other hand, is updated by minimizing a Huber loss with radius $\delta$:
\begin{equation}\label{eq:huber_dist}
    E_d = \sum_{\mathbf{u}}L_{\delta}(\mathbf{u}_d - d_i),
\end{equation}
where $\mathbf{u}$ is the assigned pixel that has a valid depth value and $\mathbf{u}_d$ is its depth. $d_i$ can be estimated by Gauss-Newton iterations. This outlier-robust mean depth not only enables the system to process low-quality depth maps but also preserves the depth discontinuity.


\subsection{Surfel Initialization}\label{initialize_surfel}

For a superpixel cluster center $C_i = [x_i, y_i, d_i, c_i, r_i]^T$ that has enough assigned pixels, we initialize one surfel $S=[S_\mathbf{p}, S_\mathbf{n}, S_c, S_w, S_r, S_t, S_i]^T$ in an outlier-robust way. The intensity $S_c$ is initialized as the mean intensity of the cluster $c_i$. $S_i$ is initialized as the index of the reference keyframe $F_{ref}$ given by the sparse SLAM system. $S_t$ is initialized as $0$ meaning that the surfel has not been fused by other frames.

The position $S_\mathbf{p}$ and normal $S_\mathbf{n}$ are initialized by using the information from all pixels of the superpixel. $S_\mathbf{n}$ is initialized as the average normal of these pixels and then fine-tuned by minimizing a fitting error defined as:
\begin{equation}\label{eq:huber_surfel}
    E_S = \sum_{\mathbf{u}}L_{\delta} ( S_\mathbf{n} \cdot ( \mathbf{p}_\mathbf{u} - \bar{\mathbf{p}} ) + b ),
\end{equation}
where $\mathbf{p}_\mathbf{u} = \mathbf{\pi}^{-1}(\mathbf{u},\mathbf{u}_d)$, $\bar{\mathbf{p}}$ is the mean of the 3D points $\mathbf{p}_\mathbf{u}$, and $b$ estimates the bias. $S_\mathbf{p}$ is defined as the point on the surfel that is observed by the camera as a pixel $[x_i, y_i]^T$:
\begin{equation}\label{eq:surfel_position_define}
\left\{
\begin{aligned}
    & S_\mathbf{n} \cdot ( S_\mathbf{p} - \bar{\mathbf{p}} ) + b  = 0\\
    &  \mathbf{\pi}(S_\mathbf{p}) = [x_i, y_i]^T
\end{aligned}
\right.
\end{equation}
and can be solved in closed-form as:
\begin{equation}\label{eq:surfel_position}
        S_\mathbf{p} = \frac{S_\mathbf{n}\cdot\bar{\mathbf{p}}-b}{S_\mathbf{n}\cdot(K^{-1}[x_i, y_i, 1]^{T})}K^{-1}[x_i, y_i, 1]^{T},
\end{equation}
where $K$ is the camera intrinsic matrix.

The surfel radius $S_r$ is initialized so that the projection of it can cover the extracted superpixel in the input intensity image:
\begin{equation}\label{eq:surfel_radius}
        S_r = \frac{S_\mathbf{p}(z) \cdot r_i \cdot \lvert\lvert K^{-1} \cdot [x_i, y_i, 1]^{T} \lvert\lvert}
                { f \cdot S_\mathbf{n} \cdot (K^{-1} \cdot [x_i, y_i, 1]^{T})},
\end{equation}
where $S_\mathbf{p}(z)$ is the depth of the surfel, and $f$ is the camera focal length. 

Most of the depth estimation methods, like stereo matching, or active stereos (e.g. Ultrastereo~\cite{ultra_stereo}) work by firstly estimating the pixel disparity $d_{dis}$ and then inverting it into depth values $d = {bf}/{d_{dis}}$, where $b$ is the baseline of the sensors. Assuming the variance of disparity estimation is $\sigma^2$, $S_w$ is initialized as the inverse variance of the estimated surfel depth:
\begin{equation}\label{eq:surfel_weight}
    S_w = \frac{b^2f^2}{S_\mathbf{p}(z)^4 \sigma^2}.
\end{equation}

\subsection{Local Map Extraction}\label{local_map}

Reconstructing large-scale environments may generate millions of surfels. However, only a subset of surfels are extracted based on the pose graph to fuse with initialized surfels due to the following reasons. Firstly, the local map fusion ensures $O(1)$ update time regardless of the reconstruction scale, and secondly, due to the accumulated tracking error of the sparse SLAM system, fusing surfels that have large drift ruins the system so that it cannot achieve global consistency even if loops are detected afterward.

Here, we introduce a novel approach that uses the pose graph from the localization system to identify local maps. With the assumption in Section~\ref{localization_system} that keyframes with the number of minimum edges to the current keyframe $F_{ref}$ below $G_{\delta}$ are locally consistent, we extract surfels attached to these keyframes as the local map. Locally consistent keyframes can be found by a breadth-first search on the pose graph. When loops are detected and edges between these keyframes are added, previous surfels can be reused so that the map growth is reduced. As shown in (d) of Fig.~\ref{fig:kitti_frame}, previous maps are reused due to the loop closure.

\subsection{Surfel Fusion}

In this section, extracted local surfels in Section.~\ref{local_map} are fused with newly initialized surfels in Section.~\ref{initialize_surfel}. Given the current camera pose estimation $\mathbf{T}_{w,c}$, the positions and normals of local surfels are firstly transformed into the current camera frame using $\mathbf{T}_{w,c}^{-1}$. Each local surfel $S^{l}$ is then back-projected into the input frame as a pixel: $\mathbf{u} = \mathbf{\pi}(S^{l}_{\mathbf{p}})$. If a surfel $S^{n}$ is initialized based on the superpixel containing $\mathbf{u}$, we determine the correspondence if they have similar depth and normals: $\lvert S^{n}_{\mathbf{p}}(z) - S^{l}_{\mathbf{p}}(z) \lvert < S^{l}_{\mathbf{p}}(z)^{2} / (bf) \cdot 2 \sigma$, and $S^{n}_{\mathbf{n}} \cdot S^{l}_{\mathbf{n}} > 0.8$. 
$S^{l}$ is fused with the corresponding surfel $S^{n}$:
\begin{equation}\label{eq:surfel_fusion_p}
\begin{aligned}
        & S^{l}_{\mathbf{p}} \leftarrow \frac{S^{l}_{\mathbf{p}}S^{l}_{w} + S^{n}_{\mathbf{p}}S^{n}_{w}}{S^{l}_{w} + S^{n}_{w}}, S^{l}_{c} \leftarrow S^{n}_{c} \\
        & S^{l}_{\mathbf{n}} \leftarrow \frac{S^{l}_{\mathbf{n}}S^{l}_{w} + S^{n}_{\mathbf{n}}S^{n}_{w}}{S^{l}_{w} + S^{n}_{w}}, S^{l}_{i} \leftarrow S^{n}_{i} \\
        & S^{l}_{t} \leftarrow S^{l}_{t} + 1, \quad S^{l}_{w} \leftarrow S^{l}_{w} + S^{n}_{w}, \\
        & S^{l}_{r} \leftarrow \min(S^{n}_{r}, S^{l}_{r}). \\
\end{aligned}
\end{equation}

After the fusion, all local surfels are transformed into the global frame using $\mathbf{T}_{w,c}$ and are moved into the global map. Surfels that are initialized in this frame but have not been fused with local maps are also transformed and added into the global map. To handle outliers, surfels with $\lvert S_{i} - F_{ref} \lvert > 10$ but are updated less than $5$ times are removed.

\section{Implementation Details}

The surfel mapping system is implemented using only CPU computing and achieves real-time performance even when it reconstructs urban-scale environments. Superpixels are initialized on the regular grid spaced $8$ pixels apart. The small-sized superpixels give the system a balance between efficiency and reconstruction accuracy. $N_s=4$, $N_c=10$ and $N_d=0.05$ are used during the pixel assignment in Equation~\ref{eq:distence_no_depth} and Equation~\ref{eq:distence_depth}. During the surfel initialization and fusion, superpixels with more than $16$ assigned pixels are used to initialize surfels. $\delta$ used in the Huber loss and the disparity error $\sigma$ are determined by the depth sensors or depth estimation methods.

\section{Experiments}

In this section, we first compare the proposed mapping system with other state-of-the-art methods using the ICL-NIUM~\cite{vafric}. The performance of the proposed system in large-scale environments is also analyzed using the KITTI dataset~\cite{kitti}. The platform to evaluate our method is a workstation with an Intel i7-7700. Finally, we use the reconstructed map to support UAV autonomous aggressive flights to demonstrate the usability of the system. In the experiments, we show that the proposed method can fuse depth maps from stereo matching, depth prediction, and monocular depth estimation.

\subsection{Reconstruction Accuracy}

We evaluate the accuracy of the reconstructed models using ICL-NIUM~\cite{vafric} and compare it with that of other mapping methods. The dataset provides rendered RGB images and the corresponding depth maps from a synthetic room. To simulate real-world data, the dataset adds noise to both RGB images and depth images. $\delta = 0.05$, $\sigma=1.0$ are used for surfel initialization. We use ORB-SLAM2 in RGB-D mode to track the camera motion. $G_\delta=20$ is used to extract the local map for fusion.

The reconstruction accuracy is defined as the mean difference between the reconstructed model and the ground truth model. Here, we compare the proposed mapping method with BundleFusion~\cite{bundlefusion}, ElasticFusion~\cite{elastic_fusion}, InfiniTAM~\cite{infiniTAM} and the recently published FlashFusion~\cite{flashfusion}. To evaluate the ability to maintain the global consistency, we also evaluate \textit{Ours w/o loop} in which the loop closure in ORB-SLAM2 is disabled.

\begin{figure}[h]
\begin{center}
\vspace{-0.2cm}
\includegraphics[width=0.8\linewidth]{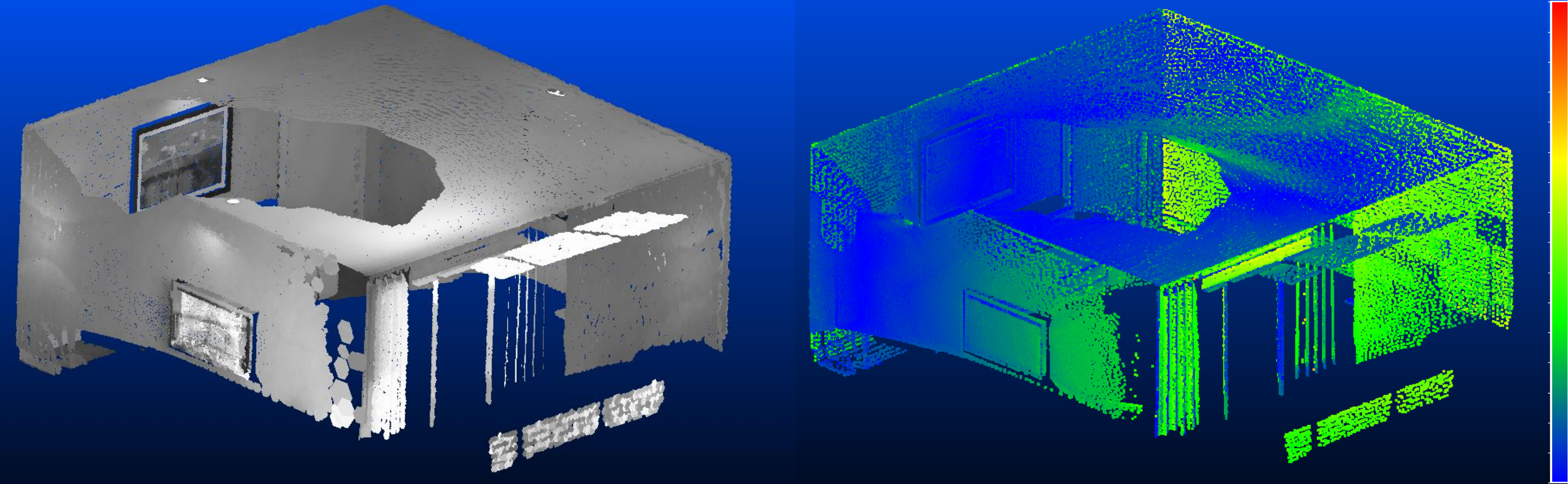}
\end{center}
\vspace*{-0.4cm}
\caption{The reconstruction result of our system on the \textit{kt3} sequence of the ICL-NIUM dataset. Left is the reconstructed meshes. Right is the error map of the surfel locations. Red represents $4$ $cm$ error and blue means $0$ $cm$ error. As visualized in the images, our method generates surfel construction that is dense and covers fine structures (such as the poles).}
\label{fig:surfel_accuracy}
\vspace{-0.5cm}
\end{figure}

The result is shown in Table~\ref{construction_table} and Fig.~\ref{fig:surfel_accuracy}. Please note that only FlashFusion~\cite{flashfusion} and our proposed system do not need GPU acceleration. BundleFusion~\cite{bundlefusion}, on the other hand, uses two high-end desktop GPU for frame reintegration and stores all the fused RGB-D frames. Although our method is designed for large-scale efficient reconstruction, it achieves similar results compared with FlashFusion. Only \textit{kt3} has global loops, and our method reduces the reconstruction error from $1.7$ to $0.8$ by removing the drift during motion tracking.

\begin{table}[h]
\centering
\caption{Reconstruction Accuracy on ICL-NIUM Dataset (cm)}
\label{construction_table}
\begin{tabular}{|l|l|l|l|l|}
\hline
Method                  & \textit{kt0} & \textit{kt1} & \textit{kt2} & \textit{kt3} \\ \hline
BundleFusion            & \textbf{0.5} & \textbf{0.6} & 0.7          & \textbf{0.8} \\ \hline
ElasticFusion           & 0.7          & 0.7          & 0.8          & 2.8          \\ \hline
InfiniTAM               & 1.3          & 1.1          & \textbf{0.1} & 2.8          \\ \hline
FlashFusion             & 0.8          & 0.8          & 1.0          & 1.3          \\ \hline
Ours                    & 0.7          & 0.9          & 1.1          & \textbf{0.8} \\ \hline
Ours w/o loop           & 0.7          & 0.9          & 1.1          & 1.7          \\ \hline
\end{tabular}
\vspace*{-0.5cm}
\end{table}

\subsection{Reconstruction Efficiency}

Most of the previous online dense reconstruction methods focus on room-scale environments using RGB-D cameras. Here, thanks to the memory and computation efficiency, we show that our method can reconstruct much larger environments, such as streets in KITTI datasets. Both the fusion update time and the memory usage are studied when the reconstruction scale grows. We use PSMNet~\cite{chang2018pyramid} to generate depth maps from stereo images and use ORB-SLAM2 in stereo mode to track the moving camera. $\delta = 0.5$, $\sigma=2.0$ are set according to the environment and the stereo method. Here, we use KITTI odometry \textit{sequences 00} for the evaluation.

\begin{figure}[h]
\begin{center}
\includegraphics[width=1.0\linewidth]{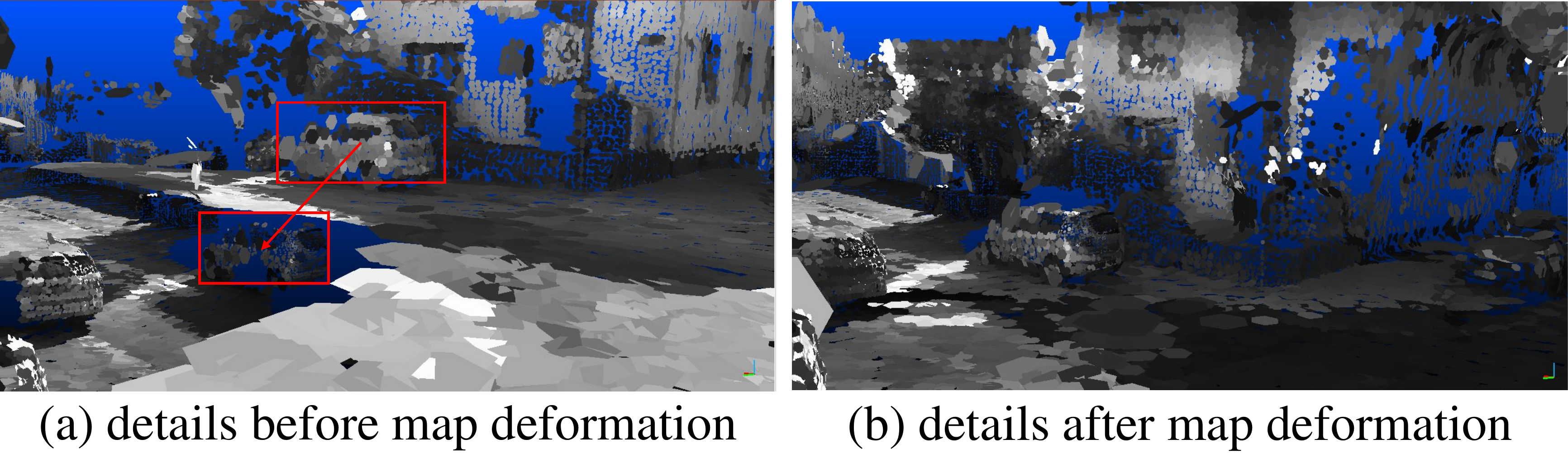}
\end{center}
\vspace*{-0.4cm}
\caption{Details of one street corner before and after the map deformation. Before the loop closure, the road and the car are misaligned due to large drift (shown in red boxes). After the loop closure and the map deformation, the drift is removed.}
\label{fig:kitti_deformation}
\vspace*{-0.6cm}
\end{figure}

\begin{figure}[h]
\begin{center}
\includegraphics[width=1.0\linewidth]{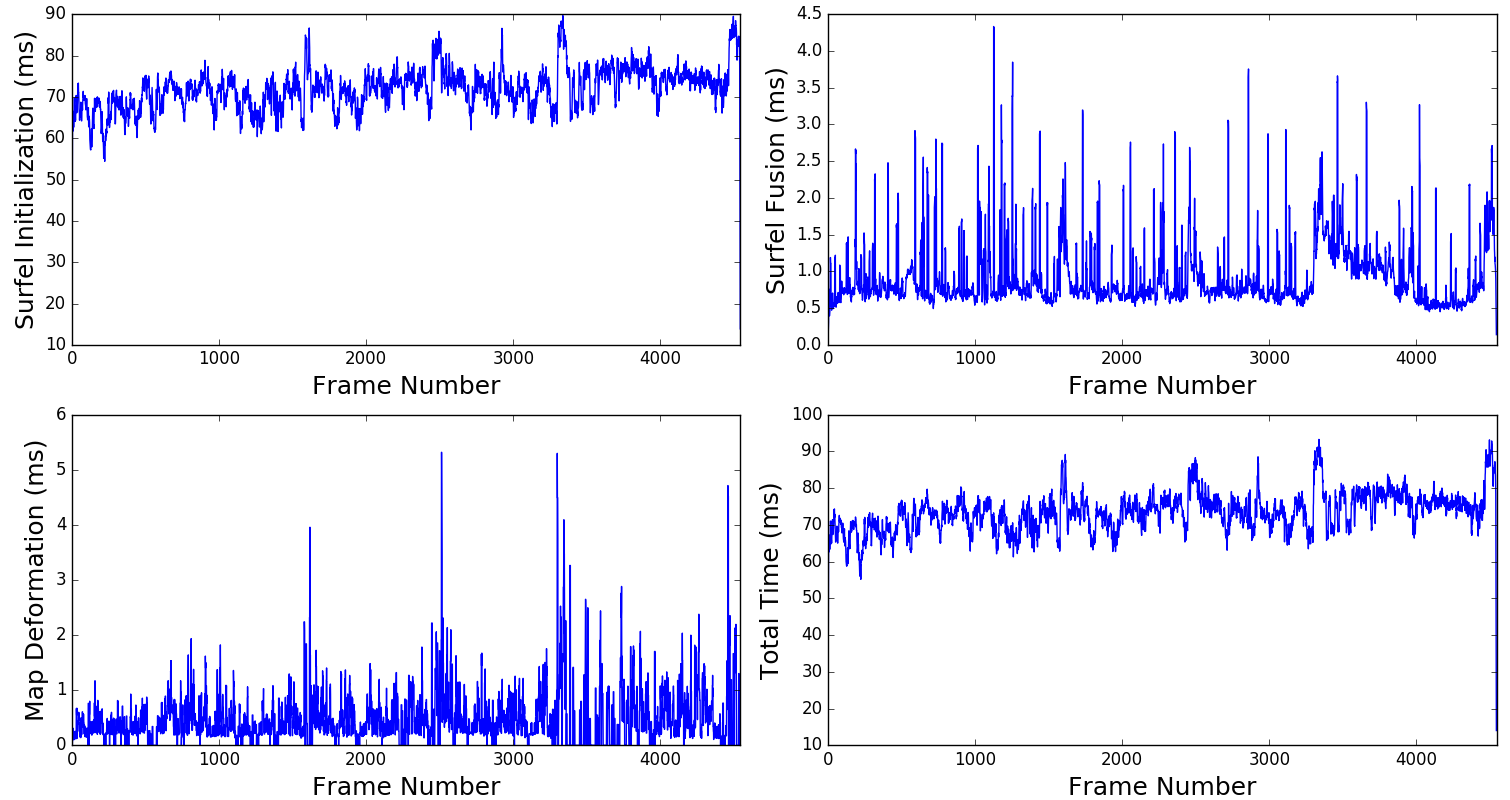}
\end{center}
\vspace*{-0.4cm}
\caption{Time efficiency of our method reconstructing KITTI odometry \textit{sequence 00}. As shown in the figure, our system achieves $10$ Hz real-time performance during the reconstruction of the KITTI sequence.}
\label{fig:kitti_time}
\vspace*{-0.6cm}
\end{figure}

\begin{figure}[h]
\begin{center}
\includegraphics[width=0.8\linewidth]{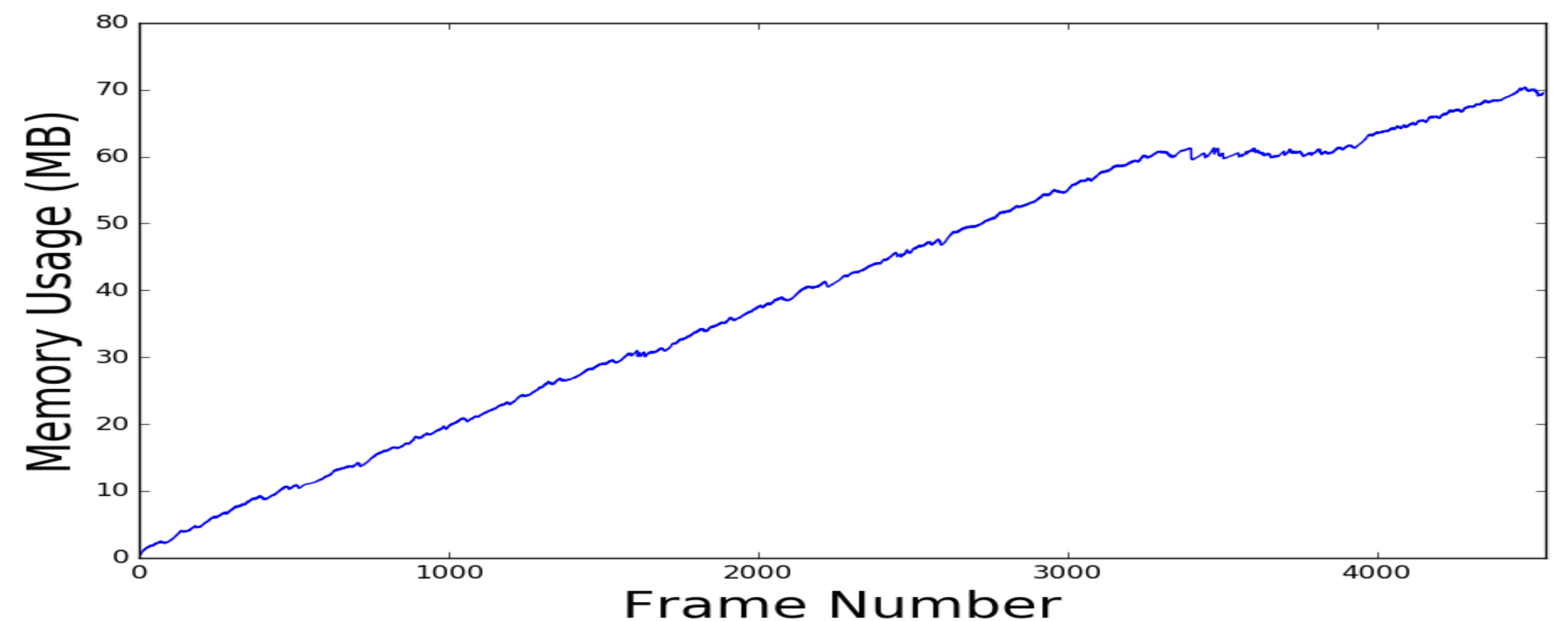}
\end{center}
\vspace*{-0.4cm}
\caption{Memory efficiency of our method reconstructing KITTI odometry \textit{sequence 00}. As shown, the memory usage grows when more frames are fused into the model. Between frame 3000 and 4000, the memory stays almost unchanged because the vehicle revisits one street and the surfels are reused.}
\label{fig:kitti_memory}
\vspace*{-0.6cm}
\end{figure}

The first row of Fig.~\ref{fig:kitti_reconstruction} shows the reconstruction result and the detail of one looped corner. Fig.~\ref{fig:kitti_deformation} shows the map before and map the map deformation. The time efficiency of our method during the KITTI \textit{sequences 00} reconstruction is shown in Fig.~\ref{fig:kitti_time}. As shown in the figure, the average fusion time is around $80$ ms per-frame, making our method more than $10$ Hz real-time using only CPU computation. Unlike other dense mapping methods, such as TSDF-based methods, our method spends most of the time extracting superpixels and initializing surfels. The outlier-robust superpixel extraction and surfel initialization enable our system to use low-quality stereo depth maps. On the other hand, the surfel fusion only consumes less than $6$ ms regardless of the environment scale. Due to the fact that ORB-SLAM2 optimizes the whole pose-graph frequently, our system deforms the map accordingly to maintain global consistency. The memory usage of the system during the runtime is shown in Fig.~\ref{fig:kitti_memory}. Between frame 3000 and 4000, the vehicle revisits one street and ORB-SLAM2 detects loop closures between the keyframes. Based on the updated pose graph, our system reuses previous surfels so that the memory grows according to the environment scale instead of the runtime.

\subsection{Using A Monocular Camera}

One of the advantages of the proposed method is that it can fuse depth maps from different kinds of sensors. In the previous sections, we showed dense mapping using rendered RGB-D images and stereo cameras. In this section, the proposed dense mapping system is used to reconstruct the KITTI sequence using only one monocular camera. Only the left images from the dataset are used to predict the depth maps~\cite{Godard_2017_CVPR}, and the camera poses are tracked using ORB-SLAM2 in RGB-D mode (with the left image and the predicted depth map). The reconstruction result is shown in the bottom row of Fig~\ref{fig:kitti_reconstruction}. During the fusion, $\sigma$ is set to $4.0$ according to the variance of the monocular depth estimation. Our method is the first one that reconstructs KITTI sequences with scale using predicted depth maps.

\subsection{Supporting Autonomous Aggressive Flights}

To prove the usability of the proposed dense mapping, we apply the system to support autonomous aggressive flights. A dense model of the environment is first built by a handheld monocular camera. Then, a flight path is generated so that the quadrotor can navigate safely and aggressively in the environment. MVDepthNet~\cite{mvdepthnet} is used to estimate monocular depth maps and VINS-MONO is used to track the camera motion. During the scene reconstruction, the proposed mapping approach corrects map drift according to the detected loops so that obstacles are consistent between different visits. We also compare the reconstruction results with CHISEL~\cite{openchisel}\footnote{https://github.com/personalrobotics/OpenChisel} using the same input images and camera poses. Since CHISEL cannot deform the map to eliminate the detected drift, fine obstacles cannot be reconstructed right when they are revisited. The results are shown in Fig.~\ref{fig:onboard_experiment}. Aggressive flights using the reconstructed maps can be found in the supplementary video. Please note that all indoor obstacles and outdoor trees are constructed accurately using our method. On the other hand, CHISEL~\cite{openchisel} cannot reconstruct fine obstacles due to the drift between different visits and the maps are not usable for autonomous flights.

\begin{figure}[h]
\begin{center}
\vspace*{-0.3cm}
\includegraphics[width=1.0\linewidth]{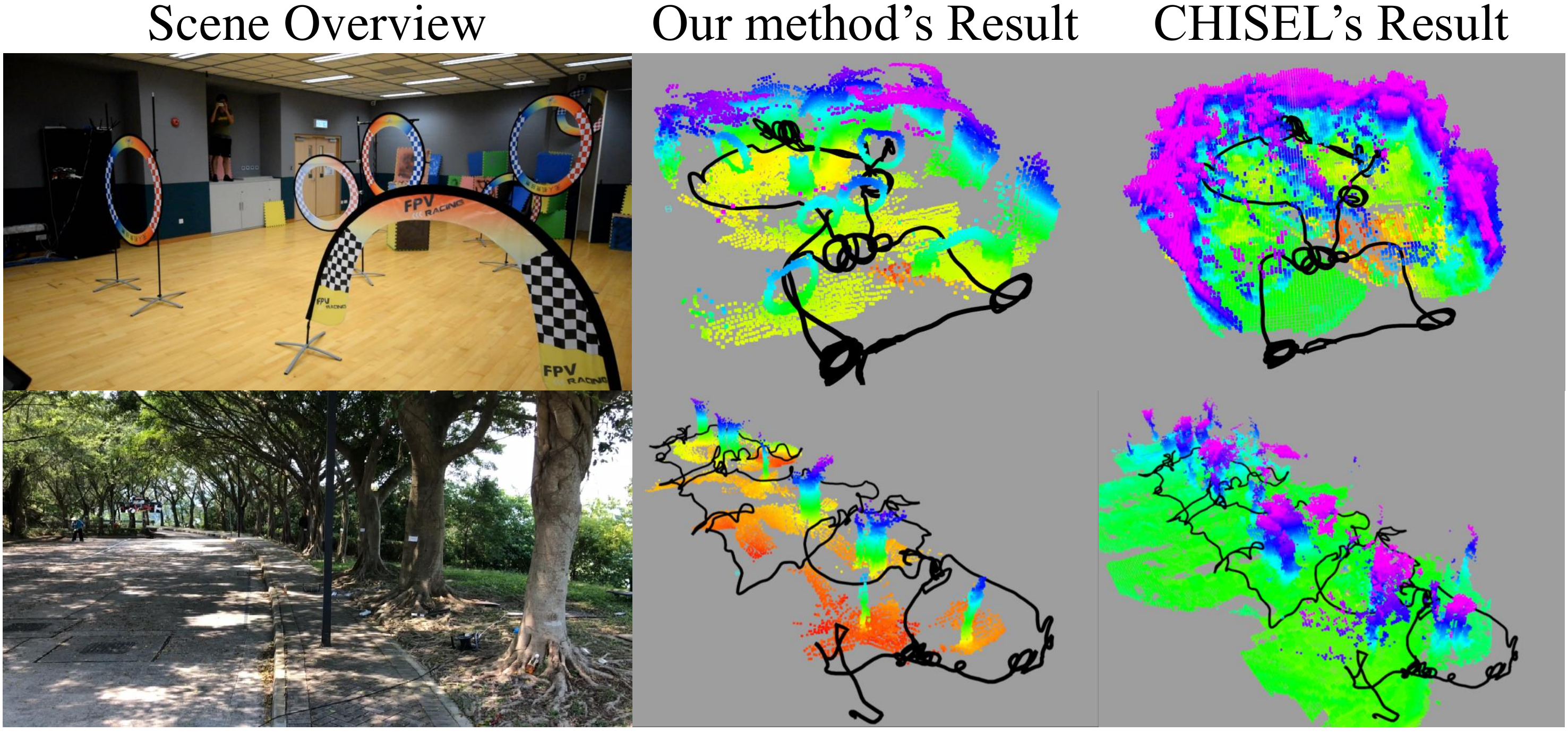}
\end{center}
\vspace*{-0.4cm}
\caption{Using the proposed method to reconstruct the environment for autonomous flights. Left is the overview of the environment. Middle is the reconstruction of our method, and right is the result of widely used CHISEL.}
\label{fig:onboard_experiment}
\vspace*{-0.4cm}
\end{figure}

\section{CONCLUSION}

In this paper, we propose a novel surfel mapping method that can fuse sequential depth maps into a globally-consistent model in real-time without GPU acceleration. The system is carefully designed so that it can handle low-quality depth maps and maintain run-time efficiency. Surfels used in our system are initialized using extracted outlier-robust superpixels. Surfels are further organized according to the pose graph of the localization system so that the system maintains $O(1)$ fusion time and can deform the map to achieve global consistency in real-time. All the characteristics of the system make the proposed mapping system suitable for robot applications.

\section{Acknowledgement}
This work was supported by the Hong Kong PhD Fellowship Scheme.

{
\bibliographystyle{unsrt}
\bibliography{template}

\begin{thebibliography}{10}

\bibitem{kitti}
A.~Geiger, P.~Lenz, and R.~Urtasun.
\newblock Are we ready for autonomous driving? the {KITTI} vision benchmark
  suite.
\newblock In {\em Conference on Computer Vision and Pattern Recognition
  (CVPR)}, 2012.

\bibitem{Godard_2017_CVPR}
C.~Godard, O.~M. Aodha, and G.~J. Brostow.
\newblock Unsupervised monocular depth estimation with left-right consistency.
\newblock In {\em Proc. of the {IEEE} Int. Conf. on Pattern Recognition}, July
  2017.

\bibitem{izadi2011kinectfusion}
S.~Izadi, D.~Kim, O.~Hilliges, D.~Molyneaux, R.~Newcombe, P.~Kohli, J.~Shotton,
  S.~Hodges, D.~Freeman, A.~Davison, and A.~Fitzgibbon.
\newblock {KinectFusion: real-time 3D reconstruction and interaction using a
  moving depth camera}.
\newblock In {\em Proceedings of the 24th annual ACM symposium on User
  interface software and technology}, Santa Barbara, CA, USA, October 2011.
  ACM.

\bibitem{tsdf}
B.~Curless and M.~Levoy.
\newblock A volumetric method for building complex models from range images.
\newblock In {\em Proceedings of the 23rd annual conference on Computer
  graphics and interactive techniques}. ACM, 1996.

\bibitem{kintinuous}
T.~Whelan, M.~Kaess, M.~Fallon, H.~Johannsson, J.~Leonard, and J.~McDonald.
\newblock {Kintinuous: Spatially extended KinectFusion}.
\newblock In {\em RSS Workshop on RGB-D: Advanced Reasoning with Depth
  Cameras}, 2014.

\bibitem{openchisel}
M.~Klingensmith, I.~Dryanovski, S.~S. Srinivasa, and J.~Xiao.
\newblock {Chisel}: Real time large scale {3D} reconstruction onboard a mobile
  device using spatially hashed signed distance fields.
\newblock In {\em Proc. of Robot.: Sci. and Syst.}, Rome, Italy, July 2015.

\bibitem{bundlefusion}
A.~Dai, M.~Niessner, M.~Zollh{\"o}fer, S.~Izadi, and C.~Theobalt.
\newblock Bundlefusion: Real-time globally consistent {3D} reconstruction using
  on-the-fly surface reintegration.
\newblock {\em ACM Transactions on Graphics (TOG)}, 36(3), 2017.

\bibitem{elastic_fusion}
T.~Whelan, S.~Leutenegger, R.~F. Salas-Moreno, B.~Glocker, and A.~J. Davison.
\newblock {ElasticFusion}: Dense {SLAM} without a pose graph.
\newblock In {\em Proc. of Robot.: Sci. and Syst.}, Rome, Italy, July 2015.

\bibitem{niessner2013real}
M.~Nie{\ss}ner, M.~Zollh{\"o}fer, S.~Izadi, and M.~Stamminger.
\newblock {Real-time 3D reconstruction at scale using voxel hashing}.
\newblock {\em ACM Transactions on Graphics (TOG)}, 32(6):169, 2013.

\bibitem{tsdf_kitti}
Ioan~Andrei B\^{a}rsan, Peidong Liu, Marc Pollefeys, and Andreas Geiger.
\newblock Robust dense mapping for large-scale dynamic environments.
\newblock In {\em Proc. of the {IEEE} Int. Conf. on Robot. and Autom.}, 2018.

\bibitem{large_scale_surfel}
X.~Fu, F.~Zhu, Q.~Wu, Y.~Sun, R.~Lu, and R.~Yang.
\newblock Real-time large-scale dense mapping with surfels.
\newblock {\em Sensors}, 18(5), 2018.

\bibitem{deformation_tsdf}
T.~Whelan, M.~Kaess, J.~J. Leonard, and J.~McDonald.
\newblock Deformation-based loop closure for large scale dense {RGB-D SLAM}.
\newblock In {\em Proc. of the {IEEE/RSJ} Int. Conf. on Intell. Robots and
  Syst.}, Tokyo, Japan, November 2013.

\bibitem{infiniTAM}
O.~K{\"a}hler, V.~A. Prisacariu, and D.~W. Murray.
\newblock Real-time large-scale dense {3D} reconstruction with loop closure.
\newblock In {\em European Conference on Computer Vision}, pages 500--516,
  2016.

\bibitem{volumetric_cpu}
F.~Steinbr\"ucker, J.~Sturm, and D.~Cremers.
\newblock Volumetric {3D} mapping in real-time on a {CPU}.
\newblock In {\em Proc. of the {IEEE} Int. Conf. on Robot. and Autom.}, Hong
  Kong, China, May 2014.

\bibitem{oleynikova2017voxblox}
H.~Oleynikova, Z.~Taylor, Ma. Fehr, R.~Siegwart, and J.~Nieto.
\newblock Voxblox: Incremental {3D} {Euclidean} signed distance fields for
  on-board mav planning.
\newblock In {\em Proc. of the {IEEE/RSJ} Int. Conf. on Intell. Robots and
  Syst.}, 2017.

\bibitem{flashfusion}
L.~Han and L.~Fang.
\newblock Flashfusion: Real-time globally consistent dense {3D} reconstruction
  using {CPU} computing.
\newblock In {\em Proc. of Robot.: Sci. and Syst.}, Pittsburgh, USA, 2018.

\bibitem{cpu_surfel}
J.~St{\"u}ckler and S.~Behnke.
\newblock Multi-resolution surfel maps for efficient dense {3D} modeling and
  tracking.
\newblock {\em Journal of Visual Communication and Image Representation},
  25(1):137--147, 2014.

\bibitem{murORB2}
Ra\'ul Mur-Artal and Juan~D. Tard\'os.
\newblock {ORB-SLAM2}: an open-source {SLAM} system for monocular, stereo and
  {RGB-D} cameras.
\newblock {\em IEEE Transactions on Robotics}, 33(5):1255--1262, 2017.

\bibitem{qin2017vins}
Tong Qin, Peiliang Li, and Shaojie Shen.
\newblock {VINS-Mono}: A robust and versatile monocular visual-inertial state
  estimator.
\newblock {\em IEEE Transactions on Robotics}, 34(4):1004--1020, 2018.

\bibitem{slic}
R.~Achanta, A.~Shaji, K.~Smith, A.~Lucchi, P.~Fua, and S.~S{\"u}sstrunk.
\newblock {SLIC} superpixels compared to state-of-the-art superpixel methods.
\newblock {\em IEEE transactions on pattern analysis and machine intelligence},
  34(11):2274--2282, 2012.

\bibitem{ultra_stereo}
S.~R. Fanello, J.~Valentin, C.~Rhemann, A.~Kowdle, V.~Tankovich, P.~Davidson,
  and S~Izadi.
\newblock Ultrastereo: Efficient learning-based matching for active stereo
  systems.
\newblock In {\em Proc. of the {IEEE} Int. Conf. on Pattern Recognition}, pages
  1063--6919. IEEE, 2017.

\bibitem{vafric}
A.~Handa, T.~Whelan, J.~MacDonald, and A.~J. Davison.
\newblock {A benchmark for RGB-D visual odometry, 3D reconstruction and SLAM}.
\newblock In {\em Proc. of the {IEEE} Int. Conf. on Robot. and Autom.}, Hong
  Kong, May 2014.

\bibitem{chang2018pyramid}
J.~Chang and Y.~Chen.
\newblock Pyramid stereo matching network.
\newblock In {\em Proc. of the {IEEE} Int. Conf. on Pattern Recognition}, pages
  5410--5418, 2018.

\bibitem{mvdepthnet}
K.~Wang and S.~Shen.
\newblock {MVDepthNet}: real-time multiview depth estimation neural network.
\newblock In {\em Proc. of the Int. Conf. on {3D} Vis.}, Sep. 2018.

\end{thebibliography}
}

\end{document}